\documentclass{article}

\PassOptionsToPackage{numbers, sort&compress}{natbib}

\usepackage[preprint]{neurips_2023}

\usepackage[utf8]{inputenc} %
\usepackage[T1]{fontenc}    %
\usepackage{booktabs}       %
\usepackage{amsfonts}       %
\usepackage{nicefrac}       %
\usepackage{microtype}      %

\usepackage{graphicx}
\usepackage{amsmath}
\usepackage{amssymb}
\usepackage{booktabs}
\usepackage[dvipsnames]{xcolor}
\usepackage{amsfonts}
\usepackage{array}
\usepackage{arydshln}
\usepackage{bbm}
\usepackage{tikzsymbols}
\usepackage{booktabs}
\usepackage{comment}
\usepackage{dblfloatfix}
\usepackage{dsfont}
\usepackage{epsfig}
\usepackage{float}
\usepackage{array}
\usepackage{mathtools,xparse}
\usepackage{dsfont}
\usepackage{graphicx,wrapfig,lipsum}
\usepackage{mathtools,nccmath}
\usepackage{multirow}
\usepackage{relsize}
\usepackage{url}
\usepackage{pifont}
\usepackage{adjustbox} %
\usepackage{caption}
\usepackage[skip=0.5ex]{subcaption}
\usepackage{tabularray}

\usepackage{hyperref}       %
\usepackage{url}            %

\usepackage{cleveref}
\crefname{section}{Sec.}{Secs.}
\Crefname{section}{Section}{Sections}
\Crefname{table}{Table}{Tables}
\crefname{table}{Tab.}{Tabs.}
\Crefname{figure}{Figure}{Tables}
\crefname{figure}{Fig.}{Tabs.}

\newcommand{\cmark}{\ding{51}}%
\newcommand{\xmark}{\ding{55}}%

\newcommand{\suppmat}{Appendix}

\newcolumntype{C}[1]{>{\centering\arraybackslash}p{#1}}

\title{Balancing the Picture: Debiasing Vision-Language Datasets with Synthetic Contrast Sets}

\author{%
  Brandon Smith\thanks{Joint first authorship. $^\dagger$Joint senior authorship.} \\
  \And
  Miguel Farinha\footnotemark[1] \\
  \And
  Siobhan Mackenzie Hall \\
  \AND
  Hannah Rose Kirk$^\dagger$ \\
  \And
  Aleksandar Shtedritski$^\dagger$ \\
  \And
  Max Bain$^\dagger$ \\
  \AND
  \normalfont{Oxford Artificial Intelligence Society, University of Oxford} \\
  \normalfont{\url{https://github.com/oxai/debias-gensynth}}
  }

\begin{document}

\maketitle

\begin{abstract}
Vision-language models are growing in popularity and public visibility to generate, edit, and caption images at scale; but their outputs can perpetuate and amplify societal biases learned during pre-training on uncurated image-text pairs from the internet. Although debiasing methods have been proposed, we argue that these measurements of \textit{model bias} lack validity due to \textit{dataset bias}. We demonstrate there are spurious correlations in COCO Captions, the most commonly used dataset for evaluating bias, between background context and the gender of people in-situ. This is problematic because commonly-used bias metrics (such as Bias@K) rely on per-gender base rates. To address this issue, we propose a novel dataset debiasing pipeline to augment the COCO dataset with synthetic, gender-balanced contrast sets, where only the gender of the subject is edited and the background is fixed. However, existing image editing methods have limitations and sometimes produce low-quality images; so, we introduce a method to automatically filter the generated images based on their similarity to real images. Using our balanced synthetic contrast sets, we benchmark bias in multiple CLIP-based models, demonstrating how metrics are skewed by imbalance in the original COCO images. Our results indicate that the proposed approach improves the validity of the evaluation, ultimately contributing to more realistic understanding of bias in vision-language models.
\end{abstract}

\section{Introduction}

Vision-Language Models (VLMs) are  rapidly advancing in capability and have witnessed a dramatic growth in public visibility: DALL-E~\cite{Ramesh2021dalle} has more than 1.5 million users creating over 2 million images a day;  the discord channel for MidJourney~\cite{midjourneyHome2023} hosts over two million members~\cite{salkowitzMidjourney2022}; and shortly after its release, Stability.AI reported that their Stable Diffusion model~\cite{Rombach2021stablediff} had over 10 million daily active users~\cite{fatundeDigital2022}. Underpinning these powerful generative models are image-text encoders like CLIP~\cite{Radford2021clip}, which are themselves used for many discriminative tasks, such as video action recognition, open set detection and segmentation, and captioning. These encoders are pre-trained on large-scale internet scraped datasets. The uncurated nature of such datasets can translate to generated images that risk inflicting a range of downstream harms on their end users and society at large -- from bias and negative stereotypes, to nudity and sexual content, or violent or graphic imagery ~\cite{Birhane2021, cherti2022reproducible}. 

In light of these issues, coupled with growing use of generative AI, it is vital to reliably benchmark the bias in VLMs, particularly in the image-text encoders. A small emerging body of work attempts to measure bias in VLMs~\cite{agarwal2021evaluating,Berg2022,chuang2023debiasing}, or to debias their feature representations~\cite{Berg2022,chuang2023debiasing}. Yet the legitimacy of this work critically depends on both a suitable evaluation metric and an evaluation dataset to accurately depict the bias in pre-trained model weights and reliably signal whether debiasing attempts have been successful. The predominant focus on model-centric debiasing methods has overshadowed two main challenges associated with datasets and metrics: (i) the common use of cropped face datasets, such as FairFace~\cite{Karkkainen2021}, fall short because excluding contextual background presents an inaccurate and unreliable assessment of bias in naturalistic images; and (ii) even if natural, open-domain images containing contextual clues are used, they are unbalanced by identity attribute representation within contexts. This is problematic because commonly-used bias metrics, such as Bias@K, are affected by the naturally-occurring distribution of images. Thus, while using contextual images is desirable, it comes at the cost of spurious correlations, affecting the reliability of bias metrics.

In this paper, we argue that these confounding factors arising from the interaction of metric choice and biased datasets paint an unreliable picture when measuring model bias in VLMs. To counter these issues, we propose a synthetic pipeline for debiasing a dataset into contrast sets balanced by identity attributes across background contexts. Our pipeline draws on the success of contrast sets in NLPs~\cite{Gardner2020} and leverages recent advances in controllable image editing and generation \cite{Brooks2022instructpix2pix}. We illustrate our approach with a focus on gender bias and define a contrast set as containing pairs of images from COCO~\cite{COCOCaptions} where each image ID has two synthetically-edited versions (one man, one woman) where the background is fixed and only the person bounding box is edited. Our paper makes three key contributions:
(1) We demonstrate spurious correlations in the COCO dataset between gender and context, and show their problematic effects when used to measure model bias (\cref{sec:measuring}); 
(2) We present the \textsc{GenSynth} dataset, built from a generative pipeline for synthetic image editing, and a filtering pipeline using KNN with real and synthetic images to control for the quality of the generated images (\cref{sec:method_gensynth}); 
(3) We benchmark state-of-the-art VLM models~\cite{Radford2021clip, wang2021actionclip, Berg2022, ilharco_gabriel_2021_5143773}; demonstrating how balanced and unbalanced versions of the COCO dataset skew the values of bias metrics (\cref{sec:benchmarking}).

Our findings demonstrate that debiasing datasets with synthetic contrast sets can avoid spurious correlations and more reliably measure model bias. While synthetically-edited data has promise in (i) preserving privacy of subjects included in vision datasets, and (ii) adding controllability to the dataset features, it also risks introducing a real-synthetic distribution shift and stacking biases of various generative models may essentialise representations of gender (see \cref{sec:limitations}). Despite these early-stage limitations, this work starts a conversation about the importance of the interaction between dataset features with bias metrics, ultimately contributing to future work that paints a more accurate and balanced picture of identity-based bias in VLMs.

\section{Related works}

\noindent\textbf{Defining Fairness and Bias.} Fairness is a complex, context-dependent concept~\cite{Mehrabi2021, Verma2018}. Here, we adopt a narrow definition where no group is advantaged or disadvantaged based on the protected attribute of gender in retrieval settings~\cite{Burns2018,friedrich2023fair}. 
 The metrics employed in this paper, \emph{Bias@K}~\cite{wang2021gender} and \emph{Skew@K},~\cite{Geyik2019} are used to assess disparity in distribution between search query results and desired outcomes.
In this work, we assume contextual activities such as \emph{dancing, skateboarding, laughing} would not have a strong gendered prior and thus the desired distribution is one where all protected attributes have equal chance of being returned in a query that does not explicitly mention gender.\footnote{In certain specific contexts, for example, pregnant or breastfeeding women, we may not necessarily want an equal distribution of masculine and feminine images to be returned, though we must be careful to not conflate biological gender and gender identity (see \cref{sec:limitations}).}

\noindent\textbf{Measuring Model Bias.}
Measuring bias in VLMs is a growing area of research. Early work measures the misclassification rates of faces into harmful categories~\cite{agarwal2021evaluating}. Several works measure outcome bias for text-to-face retrieval~\cite{Berg2022,chuang2023debiasing, seth2023dear}, though it is unclear how such measurements made on cropped face datasets generalise to real-world settings. 
For gender fairness in open-domain images, COCO Captions~\cite{COCOCaptions} is a standard benchmark for cross-modal retrieval~\cite{wang2021gender,wang2022fairclip} and image captioning~\cite{Burns2018,zhao2021understanding}. Measuring bias in generative VLMs has also been approached ~\cite{luccioni2023stable-bias}.

\noindent\textbf{Dataset Bias.} Datasets, including those used for bias evaluation, have their own biases from curation and annotation artefacts. Image datasets have been found to include imbalanced demographic representation~\cite{buolamwini2018gender,zhao2021understanding,Wang2018,de2019does,torralba2011unbiased, wang2023overcoming}, stereotypical portrayals~\cite{caliskan2017semantics,schwemmer2020diagnosing,van2016stereotyping}, or graphic, sexually-explicit and other harmful content~\cite{Birhane2021}. Similar to~\cite{meister2022gender, wang2023overcoming}, we identify spurious gender correlations in the COCO Captions dataset and further show this renders the datasets unsuitable for current bias retrieval metrics. Techniques to reduce dataset biases range from automatic~\cite{schuhmann2022laion} to manual filtering~\cite{yang2020towards} of harmful images, such as those containing nudity~\cite{schuhmann2022laion}, toxicity, or personal and identifiable information~\cite{asano2021pass}.
Yet, these filters cannot identify subtle stereotypes and spurious correlations present in open-domain images -- making it difficult to curate a wholly unbiased natural image dataset~\cite{meister2022gender}. 

\noindent\textbf{Mitigating Dataset Bias with Synthetic Data.}
Deep networks need large amounts of labeled data, prompting the creation of synthetic datasets for various computer vision tasks~\cite{johnson2017clevr, michieli2020adversarial-gta, flying-chairs, song2017semantic}. 
More recently, progress in generative models~\cite{Ramesh2021dalle, imagen, Rombach2021stablediff} has enabled methods to synthetically generate training data ~\cite{Brooks2022instructpix2pix, peebles2022gan, li2022bigdatasetgan, zhai2018classification}. 
Similarly, text-guided editing methods~\cite{Hertz2022prompt2prompt, tumanyan2022plug, Brooks2022instructpix2pix} offer scalable and controllable image editing, potentially enhancing dataset fairness and removing issues related to existing spurious correlations. 
Several works propose the use of synthetic datasets for mitigating dataset bias, such as with GANs~\cite{sattigeri2019fairness} or diffusion models~\cite{friedrich2023fair}. However, synthetic or generated data may not necessarily represent underlying distributions of marginalised groups within populations and thus still unfairly disadvantage certain groups~\cite{altman2021synthesizing, belgodere2023auditing, bhanot2021problem, lu2023machine}. To combat these risks, fairness in generative models is an area gaining popularity: StyleGan~\cite{karras2019style} has been used to edit images on a spectrum, rather than using binary categories~\cite{Hermes2022};~\cite{friedrich2023fair} use human feedback to guide diffusion models to generate diverse human images; and~\cite{Kim2022} learn to transfer age, race and gender across images. 
Similar to our work, GAN-based frameworks~\cite{ramaswamy2021fair,denton2019image} edit an \textit{existing} face dataset to equalise attributes and enforce fairness. Our work extends this approach to open-domain images, introducing an automatic filtering technique for improving the quality of edits. To our knowledge, we are the first to propose image editing of open-domain images for fairness. Our work is also inspired by the use of contrast sets in NLP~\cite{Gardner2020}, which have been used to alter data by perturbing demographics (race, age, gender) in order to improve fairness~\cite{Qian2022}. We use synthetically-generated contrast sets by augmenting both the textual and visual input to CLIP, for a more accurate evaluation of VLM bias.

\section{Measuring Gender Bias on Natural Images}
\label{sec:measuring}
While prior works make in-depth comparisons between models, and even metrics~\cite{Berg2022}, there is a dearth of research investigating whether natural image datasets, with their own biased and spurious correlations, are suitable benchmarks to measure bias in VLMs.
In this section, we investigate the extent of dataset bias from spurious correlations in COCO (\cref{sec:spurious}) and its effect on reliably measuring model bias (\cref{sec:spurious_on_bias}).

\subsection{Preliminaries}
\label{sec:preliminaries}
We first define the bias metrics and the framework used to measure model bias on image-caption data.

\textbf{Bias@K}~\cite{wang2021gender} measures the proportions of masculine and feminine images in the retrievals of a search result with a gender-neutral text query. 
For an image $I$, we define a function $g(I) = \mathrm{male}$ if there are only individuals who appear as men in the image, and $g(I) = \mathrm{female}$ if there are only individuals who appear as women.
Given a set of \emph{K} retrieved images $\mathcal{R}_K(q)$ for a query $q$, we count the images of apparent men and women as:
$$
N_{\mathrm{male}} = \sum_{I \in \mathcal{R}_K(q)} \mathds{1}[g(I) = \mathrm{male}]
\;\;\;\;\;\ \mathrm{and} \;\;\;\;\;\
N_{\mathrm{female}} = \sum_{I \in \mathcal{R}_K(q)} \mathds{1}[g(I) = \mathrm{female}].
$$
We define the gender bias metric as:
$$
\delta_{K}(q) = \begin{cases}
      0, & N_{\mathrm{male}} + N_{\mathrm{female}} = 0 \\
      \frac{N_{\mathrm{male}} - N_{\mathrm{female}}}{N_{\mathrm{male}} + N_{\mathrm{female}}}, & \text{otherwise}.
    \end{cases}
$$
For a whole query set $Q$, we define:
\begin{equation}
\mathrm{Bias@K} = \frac{1}{|Q|} \sum_{q \in Q} \delta_{K}(q).
    \label{eq:bias_at_k}
\end{equation}

\textbf{Skew@K}~\cite{Berg2022, Geyik2019} measures the difference between the desired proportion of image attributes in $\mathcal{R}_k(q)$ for the query $q$ and the actual proportion. Let the desired proportion of images with attribute label $A$ in the set of retrieved images be $p_{d, q, A} \in [0,1]$ and the actual proportion be $p_{\mathcal{R}(q), q, A} \in [0,1]$. The resulting Skew@K of $\mathcal{R}(q)$ for an attribute label $A \in \mathcal{A}$ is:

\begin{equation}
    \mathrm{Skew@K}(\mathcal{R}(q)) = \ln{\frac{p_{\mathcal{R}_{K}(q), q, A}}{p_{d, q, A}}},
\end{equation}

where the desired proportion $p_{d, q, A}$ is the actual attribute distribution over the entire dataset. A disadvantage of Skew@K is that it only measures bias with respect to a single attribute at a time and must be
aggregated to give a holistic view of the bias over all attributes. We follow~\cite{Berg2022} and take the maximum Skew@K among all attribute labels $A$ of the images for a given text query $q$:

\begin{equation}
    \mathrm{MaxSkew@K}(\mathcal{R}(q)) = \max_{A_{i} \in \mathcal{A}} \mathrm{Skew}_{A_{i}}\mathrm{@K}(\mathcal{R}(q)),
\end{equation}

which gives us the ``largest unfair advantage''~\cite{Geyik2019} belonging to images within a given attribute. In our work, a MaxSkew@K of 0 for the attribute gender and a given text query $q$ implies that men and women are equally represented in the retrieved set of $K$ images $\mathcal{R}_K(q)$. We ignore all images with undefined attribute labels (in this case gender) when measuring MaxSkew@K.

\textbf{COCO} is a dataset of 118k images with detection, segmentation and caption annotations, covering 80 distinct categories, including people~\cite{COCO2015, COCOCaptions}.
Each image has five captions written by different human annotators. COCO is commonly used to measure gender bias in VLMs in tandem with the Bias@K metric ~\cite{wang2021gender, wang2022fairclip, chuang2023debiasing}.

\subsection{Gendered Captions and Images in COCO}
\label{sec:gendered_captions}
The bias metrics defined in \cref{sec:preliminaries} require gender attribute labels for each image and gender-neutral text queries, but these are not naturally present in captioned image data such as COCO. We describe the steps to automatically label gender for images and to neutralise gender information in captions.

\paragraph{Extracting Image Gender Labels from Captions.} We assign a gender label to each COCO image, following prior work~\cite{wang2021gender}. For each image, we concatenate all five captions into a single paragraph. If the paragraph contains only feminine words and no masculine words, the image is assigned a female label, and vice versa. If the paragraph contains words from both or neither genders, it is labeled as undefined. The full list of gendered words is detailed in the~\suppmat.
Using this procedure, we implement the function $g$ in \cref{sec:preliminaries}. %
The COCO 2017 train set contains 118,287 images, of which 30,541 (25.8\%) are male, 11,781 (9.9\%) are female, and 75,965 (64.2\%) are undefined. The COCO 2017 validation set contains 5,000 images, of which 1,275 (25.5\%), are assigned male, 539 (10.8\%) female, and 3,186 (63.7\%) undefined. This procedure gives high precision in the gender-pseudo label, as any ambiguous samples are rejected. However, images may be incorrectly labeled as undefined (lower recall) due to, for example, misspelling of the gendered words in the human-annotated captions or omission of rarer gendered terms in our keyword list.

\paragraph{Constructing Gender-Neutral Captions.} We construct gender-neutral captions by replacing gendered words with neutral ones, e.g. ``man'' or ``woman'' become ``person'', and the sentence ``A \textit{man} sleeping with \textit{his} cat next to \textit{him}'' becomes ``A \textit{person} sleeping with \textit{their} car next to \textit{them}''. The full mapping of gender-neutral words and more examples of original and neutralised captions are in the~\suppmat.

\subsection{Identifying Spurious Correlations with Gender}
\label{sec:spurious}

\begin{figure*}
    \centering
    \includegraphics[width = 0.9\textwidth]{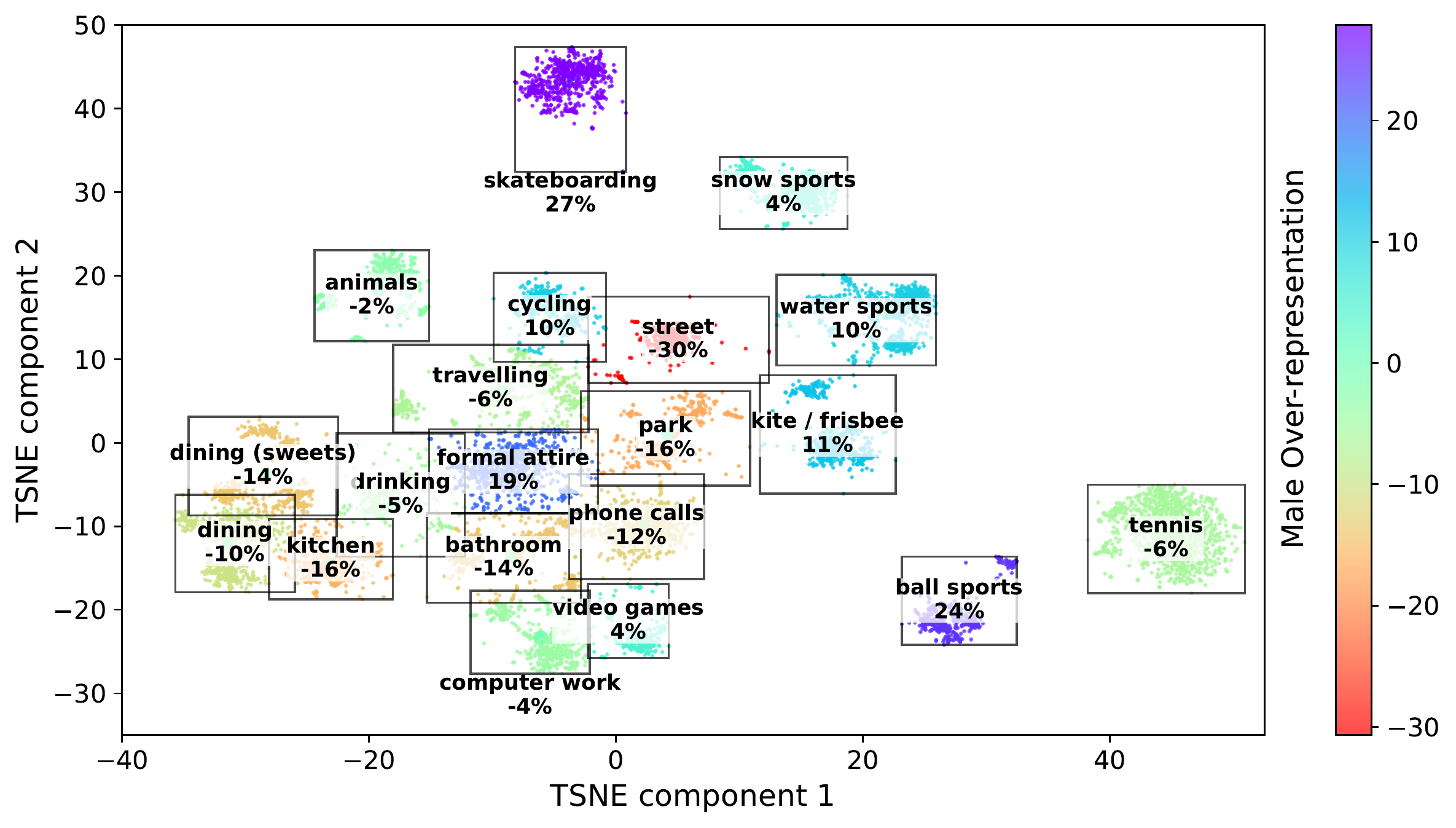}
    \caption{t-SNE clusters ($M=20$) of gender-neutralised caption embeddings. Each cluster is manually assigned a name, then coloured and labelled according to its male over-representation factor. The male over-representation factor is the difference between the percentage of male images in the particular cluster and the percentage of male images overall in the dataset.}
    \label{fig:tsne}
\end{figure*}

As reported above, COCO contains more than twice as many male images as it does female ones. This will inevitably affect retrieval-based bias metrics, as there will be more male images in the retrievals. One na\"{i}ve way to fix this is to undersample the male images in order to arrive at a \textit{Balanced} COCO dataset. 
However, ensuring equal distribution of demographic attributes does not necessarily ensure the dataset is unbiased as a whole. Spurious correlations can result in subsets of the data being highly correlated with certain attributes. Here we explore whether for certain contexts in the COCO dataset, e.g., skateboarding, one gender is over-represented. We take two approaches to evidence these spurious correlations.

\paragraph{K-means Clusters with Caption Embeddings.} First, we find semantic clusters of captions and evaluate the gender balance within them. For every image $I_n$, we embed its gender-neutralised captions $C_n^k$, where $k = \{1, \ldots , K\}$ represents the $K$ captions of the image, with RoBERTa~\cite{liu2019roberta} to get features $f_n^k$. We average the features to get $f_n = \frac{1}{K}\sum_{k=1}^{K}f_n^k$. Next, we cluster the features $f_n, n = \{1,\ldots ,N\}$ into $M=20$ clusters with K-Means. Finally, for each cluster, we extract salient words using Latent Dirichlet Allocation (LDA) and give a manually-defined cluster label. In~\cref{fig:tsne} we show a t-SNE representation of the discovered clusters, together with the degree of male over-representation. We see that in sports-related concepts men are over-represented, whereas in scenes in kitchens, bathrooms, streets, and parks, women are over-represented. For a list of all discovered classes and salient words according to LDA, refer to the~\suppmat.

\paragraph{Spurious Correlations Classifier.} Following~\cite{schwemmer2020diagnosing}, we investigate the presence of spurious correlations by training classifiers to predict binary gender labels of images and captions where the explicit gender information is removed for both training and testing. Specifically, for the image classifier (ResNet-50) we replace all person bounding boxes with black pixels; and for the caption classifier (BERT-base) we use the gender-neutralised captions. The training and testing data is COCO train and validation defined in \cref{sec:gendered_captions} but with undefined images dropped. On unseen data, the text-only classifier on gender-neutralised captions achieves 78.0\% AUC, and the image-only classifier on person-masked images achieves 63.4\% AUC. Given that a random chance model achieves 50\% AUC and an image classifier on unmasked images achieves 71.9\% AUC, it is clear that spurious background correlations in the image, as well as biases in the caption, provide a significant signal to predict gender of the person in the image even when there is no explicit gender information. %

\subsection{The Effect of Dataset Bias on Model Bias Measurement}
\label{sec:spurious_on_bias}

The dataset used for bias evaluation significantly affects the model bias measurement. This is exemplified by a theoretically fair model, which we instantiate as a TF-IDF (Term Frequency - Inverse Document Frequency) ranking model for caption-to-caption retrieval on gender-neutralised captions. Despite being based on a simple numerical statistic of word occurrences, devoid of any inherent gender bias, this model still exhibits non-zero bias when evaluated on COCO captions.
Our findings, reported in \cref{tab:spurious_metrics}, include Bias@K and MaxSkew@K measurements on COCO Val, compared against a random model and CLIP. For Balanced COCO Val, all models register an approximate Bias@K of zero, a consequence of the metric's signed nature that tends to average towards zero over many directions of spurious correlations on biased but balanced data. Yet, for unbalanced data, Bias@K shifts towards the over-represented attribute, making it an unsuitable metric for model bias measurement.
MaxSkew@K, despite being an absolute measure, is not exempt from these issues. It still records large values for the theoretically fair model and the random model, suggesting that the established framework may be inadequate for bias measurement on natural image datasets that inherently possess their own biases.

\begin{table}[]
\centering
\footnotesize
\caption{Comparison of model gender bias for CLIP~\cite{Radford2021clip}, a theoretically fair model (TF-IDF on non-gendered words) and a random model, on the COCO validation set under unbalanced and balanced (with standard deviation computed over 5 runs) settings.} %
\small
\resizebox{0.95\textwidth}{!}{
\begin{tabular}{@{}lrrrr|rrrr@{}}
\toprule
\multicolumn{1}{c}{\multirow{3}{*}{Model}} & \multicolumn{4}{c|}{COCO Val} & \multicolumn{4}{c}{COCO Val (Balanced)} \\ \cmidrule(l){2-9} 
\multicolumn{1}{c}{} & \multicolumn{2}{c}{Bias@K} & \multicolumn{2}{c|}{MaxSkew@K} & \multicolumn{2}{c}{Bias@K} & \multicolumn{2}{c}{MaxSkew@K} \\
\multicolumn{1}{c}{} & \multicolumn{1}{c}{K=5} & \multicolumn{1}{c}{K=10} & \multicolumn{1}{c}{K=25} & \multicolumn{1}{c|}{K=100} & \multicolumn{1}{c}{K=5} & \multicolumn{1}{c}{K=10} & \multicolumn{1}{c}{K=25} & \multicolumn{1}{c}{K=100} \\ \midrule
Random Model & 0.37 & 0.40 & 0.15 & 0.06 & 0.00$_{\pm 0.07}$ & 0.00$_{\pm 0.07}$ & 0.14$_{\pm 0.00}$ & 0.07$_{\pm 0.00}$ \\ 
Fair Model (TF-IDF) & 0.22 & 0.24 & 0.29 & 0.22 & -0.06$_{\pm 0.00}$ & -0.08$_{\pm 0.00}$ & 0.25$_{\pm 0.00}$ & 0.18$_{\pm 0.00}$ \\ %
CLIP & 0.20 & 0.23 & 0.28 & 0.23 & -0.03$_{\pm 0.01}$ & -0.06$_{\pm 0.01}$ & 0.24$_{\pm 0.00}$ & 0.19$_{\pm 0.01}$ \\

\bottomrule
\end{tabular}
}
\label{tab:spurious_metrics}
\end{table}

\section{\textsc{GenSynth}: A Synthetic Gender-Balanced Dataset using Contrast Sets}
\label{sec:method_gensynth}

\begin{figure}[b!]
    \centering
    \includegraphics[width=0.95\textwidth]{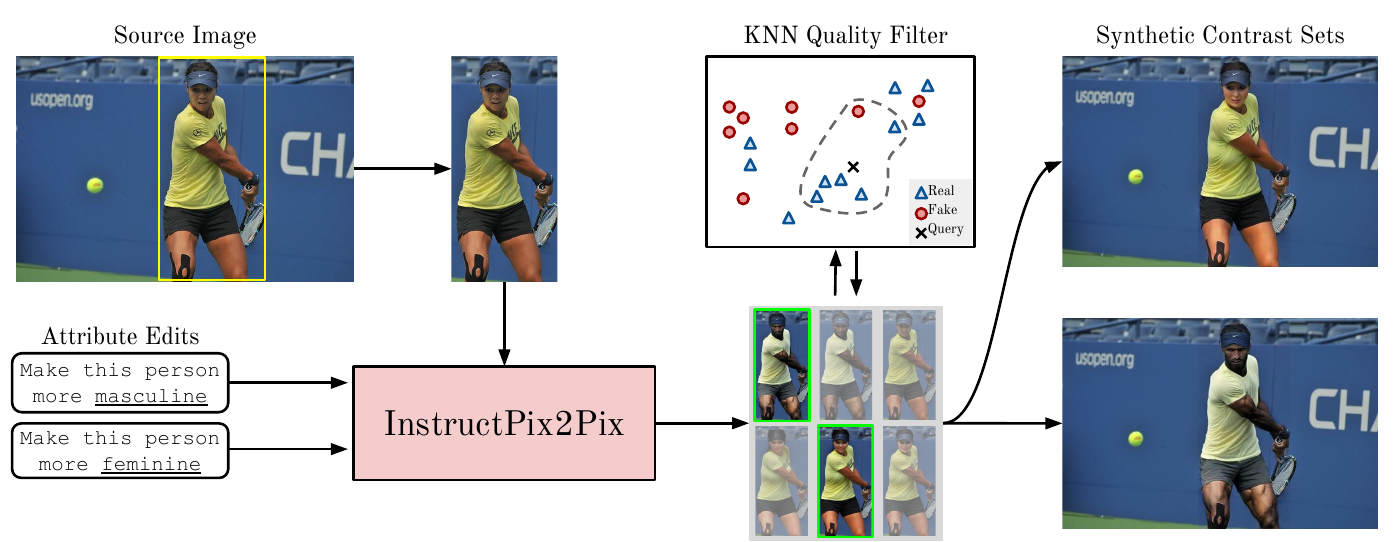}
    \caption{An overview of our pipeline for \textit{dataset debiasing} across a target attribute, in this case gender, ensuring equal demographic representation. A source image containing a person is given as input to InstructPix2Pix along with instructions to synthesise each attribute label. The resulting edits are filtered for quality via K-Nearest Neighbour (KNN) thresholding to ensure realistic-looking edits for each attribute label (male and female).}
    \label{fig:pipeline}
\end{figure}

Given the limitations of measuring Bias@K and MaxSkew@K on natural images and the spurious correlations in existing datasets, we propose a framework for editing natural images into \textit{synthetic contrast sets} that remove spurious background correlations along the attribute of interest (see \cref{fig:pipeline}), and apply the pipeline on COCO to obtain the \textsc{GenSynth} dataset (see \cref{fig:pipeline}). We first synthetically edit the person in images to cover both gender labels with fixed background context (\cref{sec:editing_images}), followed by automatic filtering that ensures the quality and correctness of the edited persons (\cref{sec:filtering}). Finally, we verify the quality of the edited images and the filtering method (\cref{sec:evaluate_pipeline}).
While we implement this for the gender attribute, in practice, our pipeline could be used to generate synthetic contrast sets for other identity attributes, requiring only the availability of person bounding boxes for the source images.

\subsection{Synthetically Editing Images}
\label{sec:editing_images}
Leveraging advancements in text-conditioned image generation and editing, we use an instruction-based model, InstructPix2Pix~\cite{Brooks2022instructpix2pix}, for editing objects in an image -- referred to as the \textit{source} image -- while keeping the background unchanged. 
We edit source images from COCO that (i) contain only one person, inferred from the number of person bounding boxes; and (ii) have a defined gender label, as defined in~\cref{sec:gendered_captions}. These restrictions remove ambiguity. Next, we crop the image to the single person bounding box and feed it to InstructPix2Pix~\cite{Brooks2022instructpix2pix} along with multiple edit instructions for each attribute label (\cref{tab: gender_sentences}). The edited person is then replaced in the source image. By only editing the appearance of the person in the image, we preserve the background content and minimize distortion -- empirically, we found editing the entire \textit{source image} rather than just the \textit{source person} produced lower quality edits with significant hallucination. For further implementation details, refer to the~\suppmat.

\begin{table}[!h]
\centering
\footnotesize
\caption{Templates used for prompt editing.}
\label{tab:templates_prompts}
\begin{tabular}{lcc}%
    \toprule 
    \multirow{2}{*}{Template} & \multicolumn{2}{c}{Instruction} \\
    & \textbf{Feminine} & \textbf{Masculine} \\
    \midrule
    Make this person more \{\}& feminine & masculine \\
    Make this person look like a \{\}& woman & man \\
    Turn this person into a \{\}& woman & man \\
    Convert this into a \{\}& woman & man\\
    \bottomrule 
\end{tabular}
\label{tab: gender_sentences}
\end{table}

\subsection{Automatic Quality Filtering of Edited Images}
\label{sec:filtering}
The synthetic edits with InstructPix2Pix~\cite{Brooks2022instructpix2pix} can often be of low quality or fail to edit the source person's attribute into the target attribute. 
In order to ensure the quality and gender accuracy of our synthetic image sets, we introduce an automatic filtering method using K-Nearest Neighbor (KNN), similar to~\cite{gu2020giqa} who use KNN to score GAN-generated images. 

First, we embed a collection of (i) source person bounding boxes, denoted as $R = \{r_1, r_2, ..., r_n\}$, and (ii) synthetically-edited person bounding boxes, denoted as $S = \{s_1, s_2, ..., s_m\}$ using CLIP.
For each synthetic box $s_i$, we identify its K-nearest neighbors in this feature space, denoted as $N_{s_i} = \mathrm{KNN}(s_i, R \cup S)$ using the Euclidean distance between the embeddings.
If the proportion of real images within $N_{s_i}$, denoted as $P_R(s_i)$, and the proportion of images corresponding to the target gender of $s_i$, denoted as $P_G(s_i)$, exceed predetermined thresholds $\tau_R$ and $\tau_G$ respectively, the edited image $s_i$ is accepted: 

\begin{equation}
P_R(s_i) = \frac{1}{K}\sum_{r \in N_{s_i}} \mathbbm{1}(r \in R) \;\;\; \mathrm{and} \;\;\;
P_G(s_i) = \frac{1}{K}\sum_{r \in N_{s_i}} \mathbbm{1}(\mathrm{gender}(r) = \mathrm{gender}(s_i)),
\end{equation}

\begin{equation}
\mathrm{accept}(s_i) = 
\begin{cases} 
1 & \text{if } P_R(s_i) > \tau_R \text{ and } P_G(s_i) > \tau_G \\
0 & \text{otherwise}.
\end{cases}
\end{equation}

This process ensures that the accepted images are of high quality and accurately reflect the target gender change. We only retain images where the entire set of edits per unique COCO ID has at least one accepted male and female edit, then randomly select one edit for each gender from images that pass the filter. For examples of edits at each decile of $\tau_R$, see the~\suppmat.

\subsection{Verifying the Quality of \textsc{GenSynth}}
\label{sec:evaluate_pipeline}
We evaluate the quality of the \textsc{GenSynth} dataset in two ways. First, to measure the correctness of the targeted gender edit, we use CLIP to zero-shot classify the gender of people in the images. Second, to measure the semantic similarity of the edited image to the caption, we measure the text-to-image retrieval performance of CLIP on the synthetic text-image captions.
For this, we edit the captions using the reverse procedure in~\cref{sec:gendered_captions} to reflect the gender of the person in the edited image. 
Then, for each image $I_i$ in \textsc{GenSynth}, where $i \in \{1,2,\ldots,N\}$, we have a set of $n$ captions $C_{i}^j$, $j \in \{1,2,\ldots,n\}$. For each caption $C_{i}^j$, we perform a retrieval operation from the COCO validation set combined with the query image $I_i$, to find a set of $K$ images that most closely match the caption, according to Euclidean distance of CLIP features. We denote this retrieved set as $R_{i}^{j}(K)$. The retrieval performance is evaluated using Recall at $K$ (R@K), which is defined as
$
R@K = \frac{1}{Nn} \sum_{i=1}^{N} \sum_{j=1}^{n} \mathbbm{1}(I_i \in R_{i}^{j}(K)).
$

We compare \textsc{GenSynth}, against (i) the original COCO 2017 dataset (train set) of natural images containing persons; and (ii) a weak gender-editing baseline -- \textsc{GenSwap}. This baseline has the same unique COCO images as in \textsc{GenSynth}, but only with edited faces -- we replace the detected face in the COCO image with a random face of the target gender from the FairFace dataset~\cite{Karkkainen2021}. Additional implementations of \textsc{GenSwap} are provided in the~\suppmat.

As shown in~\cref{tab:evaluate_gensynth}, \textsc{GenSynth} leads to very similar zero-shot classification and retrieval results to the original COCO images. The filtering step significantly improves both metrics, successfully removing bad edits. The weak baseline, \textsc{GenSwap}, consistently scores low, showing the importance of an effective editing method.  

\begin{table}[]
\centering
\footnotesize
\caption{Dataset comparison between the original COCO dataset of natural person images and synthetically edited COCO from the \textsc{GenSwap} and \textsc{GenSynth} pipelines. We report the presence of Spurious Background (BG) Correlations, Zero-Shot (ZS) Gender Accuracy, and Text-to-Image Retrieval Recall@K (R@K) amongst COCO Val 5k images using CLIP. \textit{Unfilt.} refers to the synthetic pipeline without automatic quality filtering.}
\begin{tabular}{@{}lrcccccc@{}}
\toprule
\multicolumn{1}{c}{\multirow{2}{*}{\begin{tabular}[c]{@{}c@{}}COCO-Person\\ Dataset\end{tabular}}}& \multicolumn{1}{c}{\multirow{2}{*}{\# Images}} & \multicolumn{1}{c}{\multirow{2}{*}{\begin{tabular}[c]{@{}c@{}}Edits per\\ Image\end{tabular}}} & \multirow{2}{*}{\begin{tabular}[c]{@{}c@{}}Spurious BG.\\ Correlations\end{tabular}} & \multicolumn{1}{c}{\multirow{2}{*}{\begin{tabular}[c]{@{}c@{}}ZS Gender\\ Acc. (\%) $\uparrow$\end{tabular}}} & \multicolumn{3}{c}{Text-to-Image Retrieval $\uparrow$} 
\\ \cmidrule(l){6-8} 
\multicolumn{1}{c}{} & \multicolumn{1}{c}{} &  & & \multicolumn{1}{c}{} & \multicolumn{1}{c}{R@1} & \multicolumn{1}{c}{R@5} & \multicolumn{1}{c}{R@10} \\ \midrule
Original & 11,541  & - &\cmark & 93.6 & 30.9 & 54.4 & 64.9 \\
\textsc{GenSwap} & 3,973 & 2 & \xmark & 67.9 & 19.0 & 39.8 & 50.4 \\ \midrule
\textsc{GenSynth} (unfilt.) & 11,541 & 16 & \xmark & 83.9 & 22.4 & 43.4 & 53.8 \\
\textsc{GenSynth} & 3,973 & 2 & \xmark & 95.5 & 29.2 & 52.8 & 62.8 \\ \bottomrule
\label{tab:evaluate_gensynth}
\end{tabular}
\end{table}

\section{Benchmarking Vision-Language Models on Balanced and Unbalanced Evaluation Sets}
\label{sec:benchmarking}

Here we evaluate original and debiased CLIP models on the datasets described in~\cref{sec:evaluation-datasets}. We only report MaxSkew@K results, as we showed in~\cref{sec:measuring} that Bias@K is not a reliable metric for evaluating model bias.

\subsection{Evaluation Setup}
\label{sec:evaluation-datasets}

We use the following three datasets for evaluation: \textbf{\textsc{GenSynth}} consists of 7,946 images that have been generated and filtered as discussed in~\cref{sec:method_gensynth}. It consists of 3,973 unique COCO images from the train set  (62.6\% of which were originally male), with a male and female edit for each. \textbf{COCO$_{\Strichmaxerl[1.1]}$} consists of 3,973 original (unedited) images with the same unique COCO IDs as \textsc{GenSynth}. All images contain a single person, whose gender can be identified from the caption. \textbf{COCO$_{\Strichmaxerl[1.1]\mathrm{Bal}}$} consists of 2,970 unique images from COCO$_{\Strichmaxerl[1.1]}$, randomly sampled such that there is an equal number of male and female images. We use 5 different random seeds and report average results.

We evaluate the following models: (i) the original CLIP model~\cite{Radford2021clip}; (ii) CLIP-clip~\cite{wang2021gender}, with $m=100$ clipped dimensions computed on COCO train 2017; (iii) DebiasCLIP~\cite{Berg2022}, which has been debiased on the FairFace dataset; and (iv) OpenCLIP~\cite{ilharco_gabriel_2021_5143773} models trained on LAOIN 400M and 2BN datasets~\cite{schuhmann2022laion}.
We use the ViT-B/32 variant for all models, except for DebiasCLIP, for which ViT-B/16 is used due to its availability from the authors.

\subsection{Results}
\label{sec:evaluation-results}

In~\cref{tab:results_table} we measure and compare the gender bias of CLIP-like models for the three evaluated datasets defined in~\cref{sec:evaluation-datasets}. Overall we find the MaxSkew@K metric is robust when measured on balanced (COCO$_{\Strichmaxerl[1.1]\mathrm{Bal}}$) and unbalanced data (COCO$_{\Strichmaxerl[1.1]}$), likely due to the normalization factor that considers label distribution of all the images in the dataset. CLIP-clip has the lowest gender bias across all models -- which is expected given its targeted clipping of dimensions most correlated with gender -- but comes at the cost of zero-shot image classification accuracy (60.1\% on ImageNet1k~\cite{deng2009imagenet}).
Interestingly, MaxSkew@K measured on \textsc{GenSynth} has much smaller variance between models. Given that \textsc{GenSynth} removes spurious background correlations, this suggests that a significant portion of reported model bias on natural datasets may be due to spurious correlations related to gender rather than the explicit gender of the person. %

\begin{table*}[]
\centering
\footnotesize
\caption{Comparison of Gender Bias between CLIP-like models on COCO-Person datasets. We report the MaxSkew@K in caption-to-image retrieval of gender-neutralised captions. We compare CLIP~\cite{Radford2021clip}, CLIP-clip~\cite{wang2021gender}, DebiasCLIP~\cite{Berg2022}, and OpenCLIP~\cite{ilharco_gabriel_2021_5143773} trained on LAOIN 400M \& 2BN~\cite{schuhmann2022laion}. We additionally report zero-shot image classification accuracy on ImageNet1K~\cite{deng2009imagenet}.}
\begin{tabular}{@{}clllc@{}}
\toprule
\multirow{2}{*}{\begin{tabular}[c]{@{}c@{}}COCO-Person\\ Dataset\end{tabular}} & \multirow{2}{*}{Model} & \multicolumn{2}{c}{Gender Bias $\downarrow$} & \multirow{2}{*}{\begin{tabular}[c]{@{}c@{}}ImageNet1k\\Acc. (\%) $\uparrow$\end{tabular}}\\ \cmidrule(l){3-4} 
 &  & \multicolumn{1}{c}{MaxSkew@25} & \multicolumn{1}{c}{MaxSkew@100} %
\\ \midrule
\multirow{4}{*}{COCO$_{\Strichmaxerl[1.1]}$} & CLIP & 0.27 & 0.20 & 63.2 \\
 & CLIP-clip$_{m=100}$ & 0.23 & 0.16 & 60.1 \\
 & DebiasCLIP & 0.29 & 0.22 & 67.6\\
 & OpenCLIP$_{\text{400M}}$ & 0.26 & 0.20 & 62.9 \\
 & OpenCLIP$_{\text{2B}}$ & 0.27 & 0.21 & 65.6\\
 \midrule
\multirow{4}{*}{COCO$_{\Strichmaxerl[1.1]\mathrm{Bal}}$} & CLIP & 0.26$_{\pm 0.00}$ & 0.20$_{\pm 0.00}$ & 63.2\\
 & CLIP-clip$_{m=100}$ & 0.22$_{\pm 0.00}$ & 0.15$_{\pm 0.00}$ & 60.1 \\
 & DebiasCLIP & 0.28$_{\pm 0.01}$ & 0.21$_{\pm 0.00}$ & 67.6\\
 & OpenCLIP$_{\text{400M}}$ & 0.27$_{\pm 0.00}$ & 0.20$_{\pm 0.00}$ & 62.9 \\
 & OpenCLIP$_{\text{2B}}$ & 0.27$_{\pm 0.00}$ & 0.21$_{\pm 0.00}$ & 65.6 \\ 
 \midrule
\multirow{4}{*}{\textsc{GenSynth}} & CLIP & 0.23 & 0.18 & 63.2\\
 & CLIP-clip$_{m=100}$ & 0.22 & 0.17 & 60.1 \\
 & DebiasCLIP & 0.24 & 0.19 & 67.6\\
 & OpenCLIP$_{\text{400M}}$ & 0.24 & 0.19 & 62.9\\
 & OpenCLIP$_{\text{2B}}$ & 0.23 & 0.18 & 65.6 \\
 \bottomrule
\end{tabular}
\label{tab:results_table}
\vspace{-0.5em}
\end{table*}

\section{Limitations and Ethical Considerations}
\label{sec:limitations}
\textbf{Synthetic Shifts.} By generating synthetic data, we are creating a new evaluation distribution that does not necessarily represent the real-world distribution of the respective categories. This distribution shift can also be forced in contexts where it does not necessarily make sense to either face swap or make gender edits due to factual histories or biological identity~\cite{Blodgett2021}.  %

\textbf{Assumptions of Binary Gender.} Our data relies on the binary gender labels from the COCO and FairFace datasets. COCO also presents limitations regarding race, ethnicity, and other sensitive attributes. We acknowledge this approach of using binary gender and making reference to perceived gender based on appearance oversimplifies the complexity of gender identity and biological sex, and risks erasing representation of non-binary people. Despite attempts to mitigate this limitation using terms such as ``masculine'' and ``feminine'', the resulting edits were often unusable (due to existing biases in generative models), necessitating reliance on binary and narrow terms. We advocate for future work that encodes and represents non-binary gender in datasets, and improves generalisation in generative and predictive models to non-binary terms.

\textbf{Stacking Biases.} Our pipeline uses a generative image editing model so may inadvertently introduce biases from this model via stereotypical representations of gender, e.g., if ``make this person more feminine'' over-emphasises pink clothes, or ``make this person more masculine'' over-emphasises beards. The automatic filtering step also tends to favour images with simple scene arrangements. Some model-generated images were identified as NSFW, a consequence of training on large-scale internet datasets~\cite{Birhane2021}. Future work could incorporate into our pipeline more capable and fair generative models.

\vspace{-0.5em}

\section{Conclusion}
\vspace{-0.5em}
The reliability of reported \textit{model biases} in VLMs is affected by the interaction between \textit{dataset bias} and choice of bias metric. In this paper, we demonstrated that naturalistic images from COCO have spurious correlations in image context with gender, which in turn affects how much trust can be placed in commonly-used metrics such as Bias@K: when measuring \textit{model bias}, we may in fact be measuring \textit{dataset bias}. To mitigate these problems, we proposed a pipeline for editing open-domain images at scale, creating gender-balanced contrast sets where the semantic content of the image remains the same except the person bounding box. Our method does not require manual auditing or image curation, relying instead on an effective automatic filtering method. Using this synthetically-created contrast set (\textsc{GenSynth}) we found that state-of-the-art CLIP-like models measure similarly on gender bias suggesting that measurements of model gender bias can largely be attributed to spurious model associations with gender (such as scene or background information) rather than gender itself.
Through these subsequent angles of investigation, we conclude that only focusing on model bias while ignoring how dataset artefacts affect bias metrics paints an unreliable picture of identity-based bias in VLMs. We hope our work contributes to an ongoing discussion of how to seek improved representation and diversity of identity groups in image-captioning datasets, both now and in the future.

\noindent\textbf{Acknowledgements.} This work has been supported by the Oxford Artificial Intelligence student society, the Fundação para a Ciência e Tecnologia [Ph.D. Grant 2022.12484.BD] (M.F.), the EPSRC Centre for Doctoral Training in Autonomous Intelligent Machines \& Systems [EP/S024050/1] (A.S.), and the Economic and Social Research Council Grant for Digital Social Science [ES/P000649/1] (H.R.K.). For computing resources, the authors are grateful for support from Google Cloud and the CURe Programme under Google Brain Research, as well as an AWS Responsible AI Grant.

\bibliographystyle{plainnat}
\bibliography{neurips/refs,neurips/short_strings}
\newpage
\appendix

\begin{center}
	\textbf{\Large Appendix}
\end{center}

\section{Implementation Details}
Here we provide additional implementation details about our method. 
\subsection{Gendered Words and Caption Editing}
In ~\cref{tab:gender-neutral} we show the gendered words (Masculine, Feminine) that we use for assigning each caption a gender label. Captions without either a masculine or feminine word, or captions with matches from both of these lists are labeled as \textit{undefined}. For switching or neutralising the gender in a caption, we map words across the rows of \cref{tab:gender-neutral}, so for example ``she'' could be replaced with ``he'' or ``they''. In~\cref{tab:compare_captions} we show sentences that have been gender-neutralised.

\begin{table}[!h]
\centering
\footnotesize
\setlength{\aboverulesep}{0pt}
\setlength{\belowrulesep}{0pt}
\caption{\label{tab:gender-neutral} \textbf{Gendered word pairs.} We the Masculine and Feminine words in order to classify the gender of a person in an image given its caption. When editing the gender of a caption or making it gender-neutral, we use the word from the corresponding pair for the opposite gender or the gender-neutral word, respectively.}
\begin{tabular}{lll}
    \toprule
    \textbf{Masculine} & \textbf{Feminine} & \textbf{Neutral}\\
    \midrule
    man & woman & person\\
    men & women & people\\
    male & female & person\\
    boy & girl & child\\
    boys & girls & children \\
    gentleman & lady & person\\
    father & mother & parent\\
    husband & wife & partner\\
    boyfriend & girlfriend & partner\\
    brother & sister & sibling \\
    son & daughter & child \\
    he & she & they\\
    his & hers & their\\
    him & her & them\\
    \bottomrule
\end{tabular}
\label{tab:gender_words}
\end{table}

\begin{table}[!h]
\centering
\footnotesize
\caption{\textbf{Examples of gender-neutralised captions.} We show example original COCO captions with their gender-neutralised replacements, using the corresponding words from~\cref{tab:gender-neutral}}
\begin{tabular}{p{0.45\textwidth}p{0.45\textwidth}} 
\toprule
\textbf{Original} & \textbf{Neutral} \\ 
\hline
The \textbf{\textcolor{orange}{woman}} brushes \textbf{\textcolor{orange}{her}} teeth in the bathroom. & The \textbf{\textcolor{cyan}{person}} brushes \textbf{\textcolor{cyan}{their}} teeth in the bathroom. \\
\hline
A \textbf{\textcolor{orange}{man}} sleeping with \textbf{\textcolor{orange}{his}} cat next to \textbf{\textcolor{orange}{him}}. & A \textbf{\textcolor{cyan}{person}} sleeping with \textbf{\textcolor{cyan}{their}} car next to \textbf{\textcolor{cyan}{them}}. \\
\hline
Two \textbf{\textcolor{orange}{women}} and two \textbf{\textcolor{orange}{girls}} in makeup and one is talking on a cellphone. & Two \textbf{\textcolor{cyan}{people}} and two \textbf{\textcolor{cyan}{children}} in makeup and one is talking on a cellphone. \\
\bottomrule
\end{tabular}
\label{tab:compare_captions}
\end{table}

\subsection{Image editing}
Here we provide additional details on the two image editing pipelines in the paper -- our proposed method \textsc{GenSynth}, and the weak baseline \textsc{GenSwap}.

\paragraph{\textsc{GenSynth}} We edit the COCO train set images by applying Instruct-Pix2Pix~\cite{Brooks2022instructpix2pix} on person crops (bounding boxes) with gender-editing instructions, as described in the main paper. We run Instruct-Pix2Pix for 500 denoising steps, and for each instruction, we generate an image with two text guiding scales: 9.5 and 15. We found that a smaller guiding scale sometimes does not produce the required edit, whereas too large a scale results in an image that does not look natural. Using both scales ensures there are multiple candidates for the edited image, and then we can use the filtering pipeline to discard bad edits.

\newpage
\paragraph{\textsc{GenSwap}} We use the MTCNN face detector \cite{MTCNN} to detect faces in the COCO images (for the same subset in \textsc{GenSynth}), and replace them with faces from the FairFace repository \cite{Karkkainen2021}. FairFace is a collection of face crops from the YFCC-100M dataset~\cite{Thomee2016}, labeled with gender, race and age. We only use images whose age attribute is greater than 19 and randomly sample a face crop from the target gender.

\subsection{Filtering}
For the KNN filter, we set the neighbourhood size $K = 50$,  and the thresholds $\tau_{R} = 0.08$ and $\tau_{G} = 0.5$.

\section{Spurious Correlations Analysis}
In~\cref{tab:spurious_correlations_all} we show the 20 discovered clusters using K-Means, together with the top 10 salient words according to LDA. For each cluster, we show the male-overrepresentation factor, i.e., the difference between the percentage of images in that particular cluster relative to the percentage of male images in the person class of COCO as a whole.
\begin{table}[]
\centering
\caption{\textbf{Discovered clusters in COCO Captions.}  We show all 20 clusters with their manually assigned names, together with the top 10 words according to LDA. $\Delta$M represents the deviation from gender parity for males.}
\scriptsize
\begin{tabular}{lll}
\toprule
\textbf{Name} & \textbf{Words} & \textbf{$\Delta$M} (\%)\\
\midrule

dining$_\mathrm{{drinking}}$ & wine, glass, holding, scissors, table, sitting, bottle, drinking, pouring, standing & -5.7 \\
dining$_\mathrm{{sweets}}$ & cake, banana, donut, doughnut, holding, eating, candle, table, sitting, birthday & -14.0 \\
dining$_\mathrm{{mains}}$ & pizza, eating, table, food, sandwich, sitting, holding, slice, hot, dog & -10.3 \\
sports$_\mathrm{{tennis}}$ & tennis, court, racket, ball, player, racquet, hit, holding, swinging, playing & -6.0 \\
sports$_\mathrm{{snow}}$ & ski, snow, slope, skiing, skier, snowboard, snowy, snowboarder, standing, hill & 4.7 \\
sports$_\mathrm{{skateboarding}}$ & skateboard, skate, skateboarder, riding, trick, skateboarding, ramp, young, board, child & 27.9 \\
sports$_\mathrm{{ball}}$ & baseball, bat, player, ball, soccer, field, pitch, holding, game, pitcher & 24.0 \\
sports$_\mathrm{{kite, frisbee}}$ & frisbee, kite, playing, holding, field, beach, throwing, flying, standing, child & 11.6 \\
sports$_\mathrm{{surfing}}$ & surfboard, wave, surf, surfer, riding, water, surfing, board, ocean, beach & 10.1 \\
sports$_\mathrm{{cycling, motorcycling}}$ & motorcycle, riding, bike, bicycle, street, sitting, next, standing, ride, motor & 10.5 \\
leisure$_\mathrm{{street}}$ & umbrella, holding, hydrant, standing, rain, fire, walking, street, child, black & -30.7 \\
leisure$_\mathrm{{park}}$ & sitting, dog, bench, next, holding, park, child, two, sits, frisbee & -16.9 \\
formal attire & tie, wearing, suit, standing, shirt, glass, shirt, black, white, young & 19.7 \\
computer work & laptop, sitting, computer, bed, couch, desk, room, table, using, front & -4.6 \\
animals & horse, elephant, giraffe, riding, cow, standing, sheep, next, two, brown & -2.9 \\
video games & wii, game, remote, controller, playing, video, Nintendo, holding, room, standing & 4.8 \\
kitchen & kitchen, food, standing, refrigerator, oven, cooking, counter, chef, preparing, holding & -16.2 \\
bathroom & brushing, mirror, teeth, bathroom, cat, toothbrush, taking, toilet, holding, child & -14.0 \\
travelling & standing, bear, teddy, luggage, train, next, street, bus, holding, suitcase & -6.7 \\
phone calls & phone, cell, talking, holding, sitting, cellphone, standing, looking, wearing, young & -12.8 \\
\bottomrule\\
\end{tabular}
\label{tab:spurious_correlations_all}
\end{table}

\section{Ablation Study}
We ablate the use of a CLIP vision encoder in the KNN filtering pipeline. We replace it with a DINO ViT-B/16~\cite{caron2021emerging} and repeat the analysis. We found that using DINO features is much more powerful when it comes to discriminating between the different images (real versus fake), and that the male and female images are better clustered. Accordingly, for the real vs. fake filter we use a neighborhood size of $K = 5{,}000$ and a threshold $\tau_{R} = 0.0002$  (i.e., the generated images have at least \emph{one} real neighbour). For the male vs. female filter, we use a neighborhood size of $K = 50$ and a threshold $\tau_{G} = 0.4$. We end up with 571 unique COCO images, or 1,142 images in total (with a male and female edit for each unique image). The R@K results with this dataset are 
R@1 = 33.7\%, 
R@5 = 57.1\% and
R@10 = 66.7\%,
and the zero-shot gender classification accuracy is 87.4\%. Due to the different filtering, this dataset (with DINO filtering) is smaller than \textsc{GenSynth} and the results have higher variance, but are comparable to \textsc{GenSynth}.

We evaluate MaxSkew@K on this dataset in~\cref{tab:maxskew_dino}. We observe a similar trend to the \textsc{GenSynth} dataset, where bias results across models have a smaller variance than results on the unbalanced and balanced COCO$_{\Strichmaxerl[1.1]}$ datasets. The absolute values of the bias metric are smaller, which we explain with the different images retrieved, and the variance that comes with that. 

\begin{table*}[!h]
\centering
\footnotesize
\caption{Comparison of Gender Bias between CLIP-like models on the accepted images using DINO image embeddings for KNN filtering. We report the MaxSkew@K in caption-to-image retrieval of gender-neutralised captions. We compare CLIP~\cite{Radford2021clip}, CLIP-clip~\cite{wang2021gender}, DebiasCLIP~\cite{Berg2022}, and OpenCLIP~\cite{ilharco_gabriel_2021_5143773} trained on LAOIN 400M \& 2BN~\cite{schuhmann2022laion}. We additionally report zero-shot image classification accuracy on ImageNet1K~\cite{deng2009imagenet}.}
\begin{tabular}{@{}clllc@{}}
\toprule
\multirow{2}{*}{\begin{tabular}[c]{@{}c@{}}COCO-Person\\ Dataset\end{tabular}} & \multirow{2}{*}{Model} & \multicolumn{2}{c}{Gender Bias $\downarrow$} & \multirow{2}{*}{\begin{tabular}[c]{@{}c@{}}ImageNet1k\\Acc. (\%) $\uparrow$\end{tabular}}\\ \cmidrule(l){3-4} 
 &  & \multicolumn{1}{c}{MaxSkew@25} & \multicolumn{1}{c}{MaxSkew@100} %
\\ \midrule
 
\multirow{4}{*}{\begin{tabular}[c]{@{}c@{}}\textsc{GenSynth}\\ (DINO)\end{tabular}} & CLIP & 0.15 & 0.12 & 63.2\\
 & CLIP-clip$_{m=100}$ & 0.13 & 0.10 & 60.1 \\
 & DebiasCLIP & 0.15 & 0.12 & 67.6\\
 & OpenCLIP$_{\text{400M}}$ & 0.15 & 0.12 & 62.9\\
 & OpenCLIP$_{\text{2B}}$ & 0.14 & 0.11 & 65.6 \\
 \bottomrule
\end{tabular}
\label{tab:maxskew_dino}
\vspace{-0.5em}
\end{table*} 

\section{Qualitative Dataset Examples}
\begin{figure}[h]
    \centering
    \includegraphics[width = \textwidth]{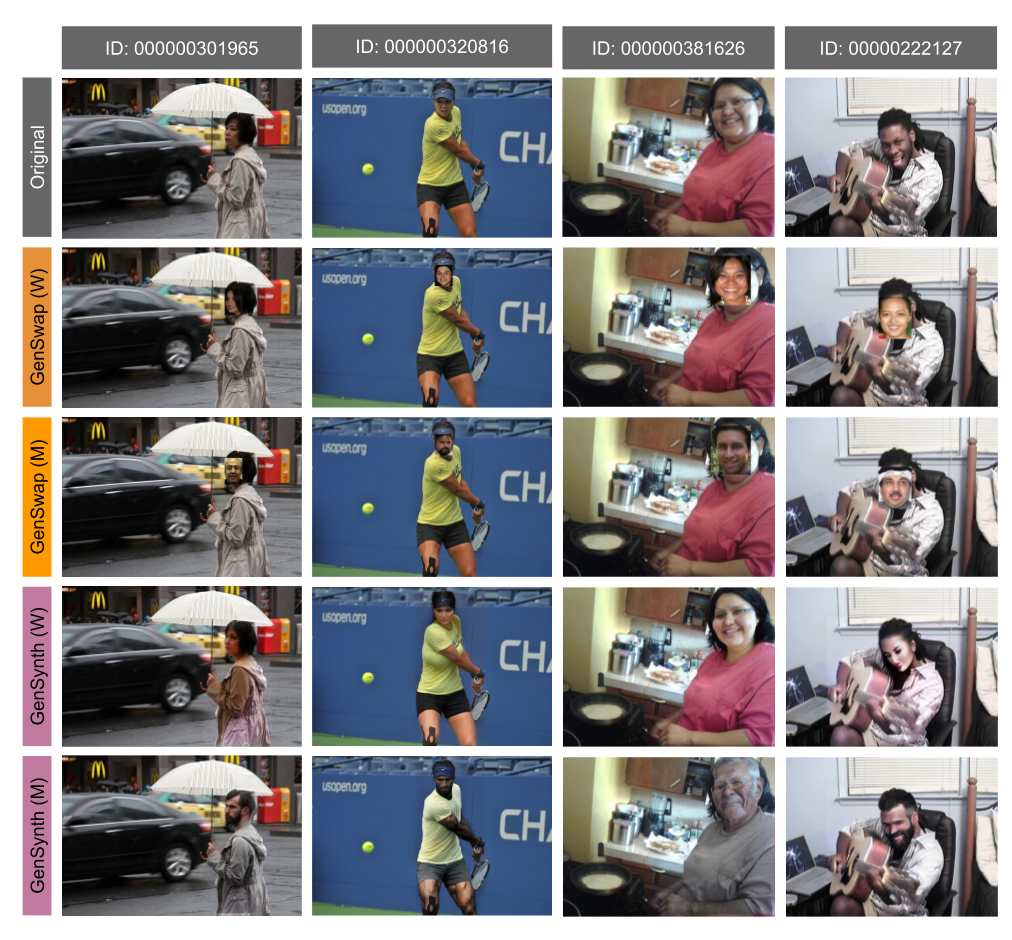}
    \caption{Randomly selected examples of \textsc{GenSynth} images showing a comparison to the original COCO image and the weak baseline \textsc{GenSwap}.}
    \label{fig:dataset_examples}
\end{figure}

In~\cref{fig:dataset_examples}, we show gender edits for the \textsc{GenSynth} and \textsc{GenSwap} datasets, alongside the original COCO image and ID. The \textsc{GenSynth} edits are more naturalistic than the \textsc{GenSwap} edits, and also make changes to the body or clothing of the subject.

\section{Comparing Image Edits Across Filtering Thresholds}
For each edited image, we calculate $P_{R}$, i.e., the ratio of real images versus fake images in the KNN clustering step. We then average $P_{R}$ for each \textit{pair} of images (the male and female edit). In~\cref{fig:knn_part_1} and~\cref{fig:knn_part_2}, we show these randomly-selected pairs of gender edits from each decile of averaged $P_{R}$ to demonstrate how our threshold filtering step improves the quality of the edited images.

\begin{figure}[!b]
    \centering
    \caption{Averaged KNN Score ($P_{R}$) for pairs of edited images using the \textsc{GenSynth} pipeline.}
    \begin{subfigure}{\textwidth}
    \includegraphics[width=\linewidth]{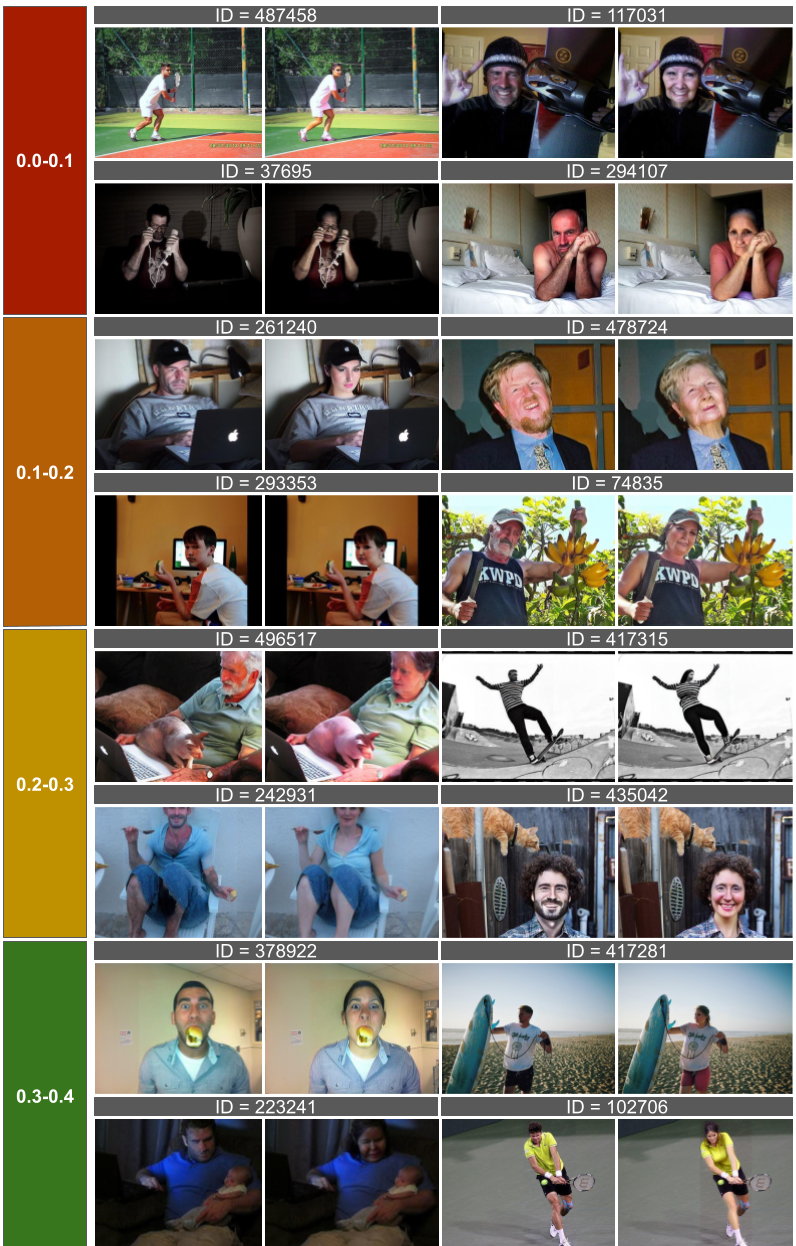}
        \subcaption{1st to 4th decile of scores.}
        \label{fig:knn_part_1}
    \end{subfigure}
\end{figure}
\begin{figure}[t]\ContinuedFloat
    \centering
    \begin{subfigure}{\textwidth}
    \includegraphics[width=\linewidth]{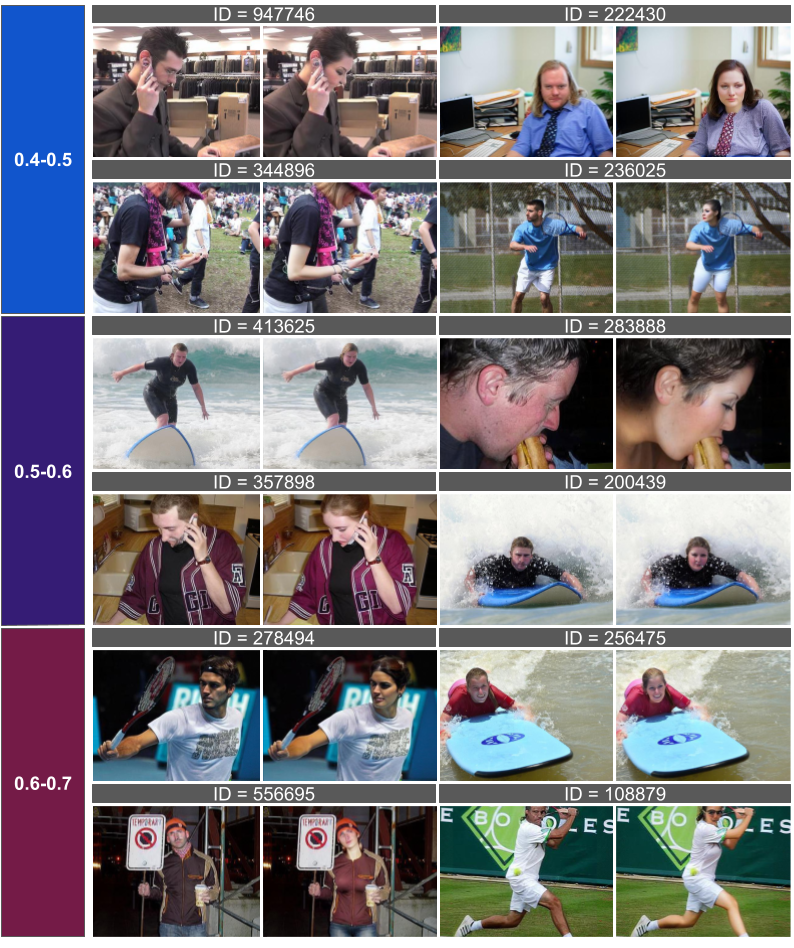}
        \subcaption{5th to 8th decile of scores. Note that there was only one image with an averaged score between 0.7-0.8, and no images in the higher deciles.}
        \label{fig:knn_part_2}
    \end{subfigure}
\end{figure}%

\end{document}